\documentclass[12pt,thmsa, a4paper]{article}%
\usepackage{amssymb}
\usepackage{amsfonts}
\usepackage{amsmath}
\usepackage{color}
\usepackage{graphicx}
\usepackage{geometry}%
\providecommand{\U}[1]{\protect\rule{.1in}{.1in}}
\newtheorem{theorem}{Theorem}

\newtheorem{fact}{Fact}
\newtheorem{definition}{Definition}

\newtheorem{proposition}[theorem]{Proposition}
\newtheorem{lemma}[theorem]{Lemma}

\newtheorem{example}{Example}

\definecolor{darkgreen}{RGB}{0,100,0}
\definecolor{darkorange}{RGB}{204,153,0}
\definecolor{amaranto}{RGB}{128,0,128}
\definecolor{paleoil}{RGB}{75,125,175}
\definecolor{oil}{RGB}{0,204,0}
\definecolor{max}{RGB}{204,0,0}
\newenvironment{boxer}
{\begin{center}
\begin{tabular}{|p{0.95\columnwidth}|}
\hline
\\
}
{
\\
\\
\hline
\end{tabular}
\end{center}
}
\newenvironment{super-boxer}
{\begin{center}
\begin{tabular}{||p{0.9\columnwidth}||}
\hline \hline \\
}
{
\\ \\ \hline \hline
\end{tabular}
\end{center}
}
\begin{document}

\title{Multialternative Neural Decision Processes}
\author{Carlo Baldassi
\and Simone Cerreia-Vioglio
\and Fabio Maccheroni
\and Massimo Marinacci
\and Marco Pirazzini}
\date{\emph{Universit\`{a} Bocconi}}
\maketitle

\begin{abstract}
We introduce an algorithmic decision process for multialternative choice that
combines binary comparisons and Markovian exploration. We show that a
preferential property, transitivity, makes it testable.\bigskip

\end{abstract}

\section{Introduction}

A decision maker aims to find the best alternative within a finite set of $A$
available ones. Had he unconstrained time (or any other relevant resource) and
were he able to make exact judgments between alternatives, he could proceed by
\emph{standard revision}. This brute force comparison-and-elimination
algorithm sequentially analyzes pairs of alternatives and permanently discards
the inferior one. After $\left\vert A\right\vert -1$ comparisons, the
incumbent solution is an optimal choice.

If time is a scarce resource and comparisons are subject to noise, say because
of the stochasticity of the information the decision maker's neural system is
able to gather, a stochastic revision procedure results. In this paper we
study a such procedure generated by an algorithmic decision process for
multialternative choice that combines binary comparisons and Markovian
exploration. We show that a preferential property, transitivity, makes it testable.

\section{The mechanics of choice}

\subsection{Kernels and transitivity}

Let $A$ be a menu of alternatives, with typical elements $i$, $j$ and $k$. We
consider a decision maker who compares alternatives $i$ and $j$ in $A$ in a
pairwise manner through a \emph{(binary) stochastic choice kernel}
$\rho:A_{\neq}^{2}\rightarrow\left[  0,1\right]  $,\footnote{The set $A_{\neq
}^{2}=\left\{  \left(  i,j\right)  \in A\times A:i\neq j\right\}  $ consists
of all distinct ordered pairs of alternatives in $A$.} where%
\[
\rho\left(  i\mid j\right)
\]
denotes the probability with which the \emph{proposal} $i$ is accepted, while
$1-\rho\left(  i\mid j\right)  $ is the probability with which the
\emph{incumbent} (or \emph{status quo}) $j$ is maintained.

\begin{definition}
A stochastic choice kernel $\rho$ is:

\begin{itemize}
\item \emph{(status-quo) unbiased} if
\[
\underset{\text{prob. of }j\text{ if status quo}}{\underbrace{1-\rho\left(
i\mid j\right)  }}=\underset{\text{prob. of }j\text{ if not status quo}%
}{\underbrace{\rho\left(  j\mid i\right)  }}%
\]
for all distinct alternatives $i\ $and $j$;

\item \emph{(strictly) positive} if $0<\rho\left(  i\mid j\right)  <1$ for all
distinct alternatives $i\ $and $j$.
\end{itemize}
\end{definition}

These properties have a simple interpretation: a stochastic choice kernel is
unbiased when it gives the incumbent alternative no special status, and
positive when it selects either alternative with a strictly positive probability.

\begin{definition}
A stochastic choice kernel $\rho$ is \emph{transitive} if
\begin{equation}
\rho\left(  j\mid i\right)  \rho\left(  k\mid j\right)  \rho\left(  i\mid
k\right)  =\rho\left(  k\mid i\right)  \rho\left(  j\mid k\right)  \rho\left(
i\mid j\right)  \label{eq:trans-ddm}%
\end{equation}
for all distinct alternatives $i$, $j$ and $k$.
\end{definition}

In words, a stochastic choice kernel is transitive when violations of
transitivity in the choices that it determines are due only to the presence of
noise. Indeed, condition (\ref{eq:trans-ddm}) amounts to require that the
intransitive cycles
\[
i\rightarrow j\rightarrow k\rightarrow i\text{\quad and\quad}i\rightarrow
k\rightarrow j\rightarrow i
\]
be equally likely.

Intransitive stochastic choice kernels may result in choices between
alternatives that feature systematic intransitivities, thus violating a basic
rationality tenet. Transitivity ensures that this is not the case.

\subsection{Binary choices and neural mechanisms}

A \emph{behavioral binary choice }(\emph{BBC})\emph{ model} is a pair of
random matrices $\left(  \mathrm{C},\mathrm{RT}\right)  $ where:

\begin{enumerate}
\item[(i)] $\mathrm{C}=\left[  \mathrm{C}_{i,j}\right]  $ consists of the
\emph{random choice variables} $\mathrm{C}_{i,j}$ taking values in $\left\{
i,j\right\}  $ that describe the outcome of the comparison between proposal
$i$ and status quo $j$;

\item[(ii)] $\mathrm{RT}=\left[  \mathrm{RT}_{i,j}\right]  $ consists of the
\emph{random response times} $\mathrm{RT}_{i,j}$ required to compare proposal
$i$ and status quo $j$, which are assumed to have finite mean and variance.
\end{enumerate}

The distributions of $\mathrm{C}$ and $\mathrm{RT}$ of a BBC model are, in
principle, both observable in choice behavior. They induce a stochastic choice
kernel $\rho_{\mathrm{C}}$ defined by
\[
\rho_{\mathrm{C}}\left(  i\mid j\right)  \doteqdot\Pr\left[  \mathrm{C}%
_{i,j}=i\right]
\]
for all proposals $i$ and incumbents $j$. Kernel $\rho_{\mathrm{C}}$ describes
the probabilistic choices that a BBC model $\left(  \mathrm{C},\mathrm{RT}%
\right)  $ induces.

A BBC model $\left(  \mathrm{C},\mathrm{RT}\right)  $ is \emph{unbiased} if
the induced $\rho_{\mathrm{C}}$ is. This is the case, for instance, when
$\mathrm{C}_{i,j}=\mathrm{C}_{j,i}$. In a similar vein, we say that a BBC
model is \emph{positive} or \emph{transitive} if the induced kernel is.
Transitivity is a preferential property of a BBC model that presupposes that a
viable choice model should not be prone to systematic errors, whatever the
underlying neural mechanism is.

Different neural mechanisms may, indeed, underlie a BBC model. A broad family
is given by the \emph{evidence threshold models}. In these models, a
stochastic process $\left\{  \mathrm{X}_{i,j}\left(  t\right)  \right\}
_{t\in\mathbb{N}}$ is given for each pair $\left(  i,j\right)  \in A_{\neq
}^{2}$ of distinct alternatives. The \emph{neural decision variable}
$\mathrm{X}_{i,j}\left(  t\right)  $ represents the evidence ---accumulated or
instantaneous--- in favor of $i$ and against $j$ that the decision maker takes
into account at time $t$. Given an evidence threshold $\beta>0$, a decision is
taken when the evidence in favor of either alternative reaches level $\beta$.
This happens at (stochastic) time
\[
\mathrm{RT}_{i,j}=\min\left\{  t\in\mathbb{N}:\left\vert \mathrm{X}%
_{i,j}\left(  t\right)  \right\vert \geq\beta\right\}
\]
and the alternative favored by evidence%
\[
\mathrm{C}_{i,j}=\left\{
\begin{array}
[c]{ll}%
i & \qquad\text{if }\mathrm{X}_{i,j}\left(  \mathrm{RT}_{i,j}\right)
\geq\beta\medskip\\
j & \qquad\text{if }\mathrm{X}_{i,j}\left(  \mathrm{RT}_{i,j}\right)
\leq-\beta
\end{array}
\right.
\]
is selected. Evidence threshold models encompass integration models, like the
drift-diffusion model and its generalizations, as well as extrema detection
models, as recently discussed by Stine, Zylberberg, Ditterich and Shadlen
(2020). This is readily seen by considering neural decision variables of the
Ornstein-Uhlenbeck form%
\[
\mathrm{X}_{i,j}\left(  t+1\right)  -\mathrm{X}_{i,j}\left(  t\right)
=-\lambda\mathrm{X}_{i,j}\left(  t\right)  +\left(  v_{i}-v_{j}\right)
\mu\left(  t\right)  +\sigma\varepsilon\left(  t\right)
\]
where $\lambda\in\left(  0,1\right)  $ captures evidence deterioration,
$\left(  v_{i}-v_{j}\right)  \mu\left(  t\right)  $ is the instantaneous
strength of the evidence in favor of $i$ over $j$,\footnote{The dependence of
$\mu$ on $t$ allows to incorporate the presence of urgency signals.} and
$\sigma$ is the standard deviation of a Gaussian white noise process
$\varepsilon$.

\section{Neural Metropolis algorithm}

We now describe an algorithmic decision process that the neural system of a
decision maker might implement when facing a multialternative menu $A$. The
process consists of a sequence of pairwise comparisons conducted via a BBC
model, the contestants of which are selected by a Markovian mechanism a la
Metropolis et al. (1953).

The algorithm starts according to an initial distribution $\mu\in\Delta\left(
A\right)  $ that describes the \textquotedblleft first
fixation\textquotedblright\ of the decision maker,\footnote{As usual,
$\Delta\left(  A\right)  $ is the set of all probability distributions on $A$,
and $\Delta_{+}\left(  A\right)  $ is its subset consisting of all probability
distributions with full support.} and proceeds through an exploration matrix
$Q$ that describes how his neural system navigates the landscape of
alternatives, as suggested by eye-tracking evidence.\footnote{See Russo and
Rosen (1975) and the more recent Krajbich and Rangel (2011) and Reutskaja et
al. (2011).} Pairs of alternatives are compared via a BBC model $\left(
\mathrm{C},\mathrm{RT}\right)  $ and the last incumbent is maintained.

The algorithm terminates according to a (stochastic) stopping time $N$.

\begin{center}
\rule{18cm}{0.04cm}

\textbf{Neural Metropolis Algorithm}

\rule{18cm}{0.04cm}
\end{center}

\noindent\textbf{Input:}$\ $\emph{Given a stopping time }$N>0$\emph{.}\medskip

\noindent\textbf{Start: }\emph{Draw }$i_{0}\ $\emph{from}$\emph{\ }A$\emph{
according to }$\mu$ \emph{and}$\medskip$

$\bullet$ \emph{set }$t_{0}=0$\emph{,\medskip}

$\bullet$\emph{ set }$j_{0}=i_{0}$\emph{.\medskip}

\noindent\textbf{Repeat: }\emph{Draw }$i_{n+1}\ \emph{from\ }A$\emph{
according to }$Q\left(  \cdot\mid j_{n}\right)  $\emph{ and compare it to
}$j_{n}$\emph{:\medskip}

$\bullet$ \emph{set }$t_{n+1}=t_{n}+\mathrm{RT}_{i_{n+1},j_{n}}$%
\emph{,\medskip}

$\bullet$\emph{ set }$j_{n+1}=\mathrm{C}_{i_{n+1},j_{n}}$\emph{;\medskip}

\noindent\textbf{until }$n+1=N$\emph{.}\medskip

\noindent\textbf{Stop: }\emph{Set }$k=j_{n}$\emph{.\medskip}

\noindent\textbf{Output: }\emph{Choose }$k$\emph{ from }$A$\emph{.}

\begin{center}
\rule{18cm}{0.04cm}
\end{center}

When the underlying BBC Model is the Drift Diffusion Model of Ratcliff (1978)
and the stopping rule corresponding to $N$ is given by a fixed deadline $T$,
the Neural Metropolis Algorithm coincides with the Metropolis-DDM of Baldassi
et al. (2020). Based on the latter paper, and independently of this one,
Valkanova (2020) studied a Markovian model of decision-making with a geometric
stopping time; but the model only considers \textquotedblleft
timeless\textquotedblright\ choice probabilities.

The Neural Metropolis Algorithm generates a Markov chain of incumbents%
\[
\mathcal{M}=\left\{  J_{0},J_{1},...\right\}
\]
with $\Pr\left[  J_{0}=j\right]  =\mu\left(  j\right)  $ for all alternatives
$j\in A$ and
\[
\Pr\left[  J_{n+1}=i\mid J_{n}=j\right]  =\underset{\text{prob. }i\text{ is
proposed}}{\underbrace{Q\left(  i\mid j\right)  }}\times\underset{\text{prob.
that }i\text{ is accepted}}{\underbrace{\rho_{\mathrm{C}}\left(  i\mid
j\right)  }}\doteqdot M\left(  i\mid j\right)
\]
for all distinct alternatives $i$ and $j$ in $A$. Thus, $M$ is the transition
matrix of this Markov chain.

The first result of this note is that multialternative choice distributions
and average decision times of the algorithm can be written in explicit form.
To this end, note that, if the incumbent at some iteration is $j$ and the
proposal is $i$, the average duration of that iteration is $\overline
{\mathrm{RT}}_{i,j}$, but then \emph{conditional on }$j$\emph{ being the
incumbent}, the average duration of an iteration is
\[
\tau_{j}=%
{\displaystyle\sum\limits_{i\in A}}
Q\left(  i\mid j\right)  \overline{\mathrm{RT}}_{i,j}%
\]
Denote by $\tau$ is the vector of those conditional expected durations.

\begin{lemma}
\label{lem:comp}The Neural Metropolis Algorithm generates the following choice
probabilities from $A$%
\begin{equation}
p_{N}=\left(
{\displaystyle\sum\limits_{m=1}^{\infty}}
\Pr\left[  N=m\right]  M^{m-1}\right)  \mu\label{eq:uella}%
\end{equation}
and the mean decision time is%
\begin{align}
\bar{T}_{N}  &  =\tau^{\intercal}\left(
{\displaystyle\sum\limits_{m=1}^{\infty}}
\Pr\left[  N=m\right]  \left(
{\displaystyle\sum\limits_{n=1}^{m}}
M^{n-1}\right)  \right)  \mu\label{eq:uesta}\\
&  =\tau^{\intercal}\left(
{\displaystyle\sum\limits_{n=1}^{\infty}}
\Pr\left[  N\geq n\right]  M^{n-1}\right)  \mu\label{eq:altra}%
\end{align}
provided $N$, $\mathrm{C}$, and $\mathrm{RT}$ are independent.
\end{lemma}

Note that denoting by $p_{ij}^{m-1}\in\left[  0,1\right]  $ the generic entry
of $M^{m-1}$, then each entry%
\[%
{\displaystyle\sum\limits_{m=1}^{\infty}}
\Pr\left[  N=m\right]  p_{ij}^{m-1}%
\]
of $%
{\displaystyle\sum\limits_{m=1}^{\infty}}
\Pr\left[  N=m\right]  M^{m-1}$ is a converging series (the average of the
function $m\mapsto p_{ij}^{m-1}$ with respect to the distribution of $N$), so
$%
{\displaystyle\sum\limits_{m=1}^{\infty}}
\Pr\left[  N=m\right]  M^{m-1}$ is a \emph{bona fide} stochastic matrix and
$p_{N}$ a \emph{bona fide} probability vector. Moreover, each entry%
\[%
{\displaystyle\sum\limits_{n=1}^{\infty}}
\Pr\left[  N\geq n\right]  p_{ij}^{n-1}%
\]
of $%
{\displaystyle\sum\limits_{n=1}^{\infty}}
\Pr\left[  N\geq n\right]  M^{n-1}$ is bounded by $%
{\displaystyle\sum\limits_{n=1}^{\infty}}
\Pr\left[  N\geq n\right]  =\mathbb{E}\left[  N\right]  $, thus the
convergence of $%
{\displaystyle\sum\limits_{n=1}^{\infty}}
\Pr\left[  N\geq n\right]  M^{n-1}$ is guaranteed if $N$ has finite
expectation. When $M$ is reversible (see next section) the previous
computation is standard because $M$ can be diagonalized,
\[
M=U\left[
\begin{array}
[c]{cccc}%
\lambda_{1} &  &  & \\
& \lambda_{2} &  & \\
&  & \ddots & \\
&  &  & \lambda_{\left\vert A\right\vert }%
\end{array}
\right]  U^{-1}%
\]
where the columns of $U$ form a basis of eigenvectors and hence
\[%
{\displaystyle\sum\limits_{m=1}^{\infty}}
\Pr\left[  N=m\right]  M^{m-1}=U\left[
\begin{array}
[c]{ccc}%
{\displaystyle\sum\limits_{m=1}^{\infty}}
\Pr\left[  N=m\right]  \lambda_{1}^{m-1} &  & \\
& \ddots & \\
&  &
{\displaystyle\sum\limits_{m=1}^{\infty}}
\Pr\left[  N=m\right]  \lambda_{\left\vert A\right\vert }^{m-1}%
\end{array}
\right]  U^{-1}%
\]
this gives even more explicit expressions for equations (\ref{eq:uella}),
(\ref{eq:uesta}), and (\ref{eq:altra}).

\begin{example}
But also in the absence of reversibility, since $%
{\displaystyle\sum\limits_{m=1}^{\infty}}
\Pr\left[  N=m\right]  M^{m-1}$ is a power series, some computation is
possible. The geometric and Poisson cases are iconic:

\begin{itemize}
\item If stopping is geometric with continuation probability $\zeta$, then%
\[%
{\displaystyle\sum\limits_{m=1}^{\infty}}
\Pr\left[  N=m\right]  M^{m-1}=\left(  1-\zeta\right)  \left(  I-\zeta
M\right)  ^{-1}%
\]
this was first proven by Valkanova (2020).

\item Instead, if stopping is Poisson with mean $\lambda$, then%
\[%
{\displaystyle\sum\limits_{m=1}^{\infty}}
\Pr\left[  N=m\right]  M^{m-1}=e^{-\lambda}e^{\lambda M}%
\]
this result is new.
\end{itemize}
\end{example}

\section{Transitive algorithms}

How can we test whether incumbents are stochastically generated by a neural
Metropolis algorithm? The next Proposition \ref{prop:trans} shows that
transitivity makes it possible to address this key question, but we need a
simple lemma before getting to it. Recall that an exploration matrix is
\emph{nice} if and only if it is symmetric and strictly positive off the diagonal.

\begin{lemma}
\label{lem:full}If the BBC is positive, and the exploration matrix $Q$ is
nice, then the transition matrix $M$ has strictly positive entries.
\end{lemma}

We are now ready to state the second result of this note.

\begin{proposition}
\label{prop:trans}The following conditions are equivalent for a positive BBC model:

\begin{enumerate}
\item[(i)] the transition matrix $M$ is reversible, for every/some nice
exploration matrix $Q$;

\item[(ii)] the stochastic choice kernel $\rho_{\mathrm{C}}$ is transitive;

\item[(iii)] there exist a probability $\pi\in\Delta_{+}\left(  A\right)  $
and a symmetric function $s:A_{\neq}^{2}\rightarrow\left(  0,\infty\right)  $
such that%
\begin{equation}
\rho_{\mathrm{C}}\left(  i\mid j\right)  =s\left(  i,j\right)  \dfrac
{\pi\left(  i\right)  }{\pi\left(  i\right)  +\pi\left(  j\right)  }%
\qquad\forall i\neq j \label{eq:Haste}%
\end{equation}

\end{enumerate}

In this case, $\pi$ and $s$ are unique. Moreover,

\begin{enumerate}
\item $\pi$ is the only element of $\Delta\left(  A\right)  $ under which $M$
is reversible and, given any alternative $i$,%
\begin{equation}
\pi\left(  j\right)  =\frac{\dfrac{\rho\left(  j\mid i\right)  }{\rho\left(
i\mid j\right)  }}{%
{\displaystyle\sum\limits_{k\in A}}
\dfrac{\rho\left(  k\mid i\right)  }{\rho\left(  i\mid k\right)  }}%
\qquad\forall j \label{eq:expli}%
\end{equation}

\item $\rho$ is unbiased if and only if $s$ is constant to $1$, that is,%
\[
\rho_{\mathrm{C}}\left(  i\mid j\right)  =\dfrac{\pi\left(  i\right)  }%
{\pi\left(  i\right)  +\pi\left(  j\right)  }\qquad\forall i\neq j
\]

\end{enumerate}
\end{proposition}

The importance of this proposition, so of transitivity, does not reside in the
existence of a unique stationary distribution $\pi$ for $M$. Indeed, any
neural Metropolis algorithm featuring a positive BBC has an irreducible and
aperiodic transition matrix, hence a unique stationary distribution $\pi$ that
may be used to approximate the frequency of incumbents that the algorithm
generates (assuming its convergence, i.e., a $T$ large enough relative to the
BBC average response times; see the next section).

Instead, the central feature of our result is that, when the BBC is\emph{
transitive}, the stationary distribution is \emph{independent }of $Q$ and can
be expressed \emph{solely }in terms of the BBC kernel $\rho_{\mathrm{C}}$.
Therefore, by knowing the observable elements of a transitive BBC and assuming
convergence of the algorithm, we can test the neural Metropolis algorithm.

A functional property like transitivity, which we claim that any viable
decision process should feature, thus characterizes a class of testable
multialternative neural decision processes.

\section{Asymptotic heuristics}

As observed above we have the following:

\begin{fact}
For each positive BBC model, and each nice exploration matrix $Q$, the
transition matrix $M$ has a unique stationary distribution $\pi$.
\end{fact}

\begin{boxer}
\textbf{What is the relation between the stationary distribution }$\pi
$\textbf{ and the algorithm's output, when }$N$\textbf{ is given by a fixed
deadline }$T$\textbf{?} Heuristically, as $T$ increases and so does the number
$n$ of iterations performed, the fraction of clock-time in which $j$ is the
incumbent and $i$ is the proposal is directly proportional to:

\begin{itemize}
\item the probability of $j$ being the incumbent and $i$ the proposal, which
is
\[
\underset{\text{prob. of }j\text{ after }n\text{ iter.s}}{\underbrace{\left(
M^{n}\mu\right)  \left(  j\right)  }}\times\underset{\text{prob. of proposal
}i\text{ given }j}{\underbrace{Q\left(  i\mid j\right)  }}\rightarrow
\pi\left(  j\right)  Q\left(  i\mid j\right)
\]

\item the average clock-time it takes to compare $i$ and $j$, which is
$\overline{\mathrm{RT}}_{i,j}$.
\end{itemize}

Assuming $Q$ is null on the diagonal (i.e., that incumbents cannot be
re-proposed), \emph{a reasonable conjecture} is that, as $T$ increases, the
total probability of $j$ being the incumbent approaches%
\[
\pi^{\ast}\left(  j\right)  \underset{\text{normalizing constant}%
}{=\underbrace{\frac{\sum\limits_{i\in A\setminus j}\pi\left(  j\right)
Q\left(  i\mid j\right)  \overline{\mathrm{RT}}_{i,j}}{\sum\limits_{k\in
A}\left(  \sum\limits_{i\in A\setminus k}\pi\left(  k\right)  Q\left(  i\mid
k\right)  \overline{\mathrm{RT}}_{i,k}\right)  }}}\qquad\forall j\in A
\]

\end{boxer}

Prliminary investigations, and numerical simulations for the very general
Ornstein-Uhlenbeck family of BBC models,\footnote{See, Busemeyer and Townsend
(1993), Bogacz et al. (2006), and Stine, Zylberberg, Ditterich and Shadlen
(2020).} provide preliminary support to the conjecture.\footnote{Python
scripts are available upon request.}

\section{Proofs and related material}

\subsection{Proof of Lemma \ref{lem:comp}}

The algorithm does not stop at iteration $0$ and the $0$-th iteration is
instantaneous. Moreover, if it stops at iteration $m$ it chooses the incumbent
$j_{m-1}$.

By independence, for each $i$, the probability of choosing $i$
\emph{conditional on stopping at iteration }$m$ is the $i$-th component
$\left(  M^{m-1}\mu\right)  _{i}$ of the vector $M^{m-1}\mu$. The probability
of stopping at iteration $m$ is $\Pr\left[  N=m\right]  $, and so the
probability of of choosing $i$ is%
\[%
{\displaystyle\sum\limits_{m=1}^{\infty}}
\Pr\left[  N=m\right]  \left(  M^{m-1}\mu\right)  _{i}%
\]
but then the choice probabilities are given by
\begin{align*}%
{\displaystyle\sum\limits_{m=1}^{\infty}}
\Pr\left[  N=m\right]  M^{m-1}\mu &  =\lim_{l\rightarrow\infty}%
{\displaystyle\sum\limits_{m=1}^{l}}
\Pr\left[  N=m\right]  M^{m-1}\mu=\lim_{l\rightarrow\infty}\left(
{\displaystyle\sum\limits_{m=1}^{l}}
\Pr\left[  N=m\right]  M^{m-1}\right)  \mu\\
&  =\left(
{\displaystyle\sum\limits_{m=1}^{\infty}}
\Pr\left[  N=m\right]  M^{m-1}\right)  \mu
\end{align*}
proving equation (\ref{eq:uella}).

By independence, \emph{conditional on }$j$\emph{ being the incumbent}, the
average duration of an iteration is
\[
\tau_{j}=%
{\displaystyle\sum\limits_{i\in A}}
Q\left(  i\mid j\right)  \overline{\mathrm{RT}}_{i,j}%
\]
Now the probability with which $j$ is an incumbent at iteration $n$ is
$\left(  M^{n-1}\mu\right)  _{j}$, then the average duration of iteration $n$
(if it takes place) is%
\[%
{\displaystyle\sum\limits_{j\in A}}
\tau_{j}\left(  M^{n-1}\mu\right)  _{j}=\tau^{\intercal}M^{n-1}\mu
\]
By independence, the average processing time, \emph{conditional on performing
}$m$ iterations is%
\[%
{\displaystyle\sum\limits_{n=1}^{m}}
\tau^{\intercal}M^{n-1}\mu=\tau^{\intercal}\left(
{\displaystyle\sum\limits_{n=1}^{m}}
M^{n-1}\right)  \mu
\]
But the probability of performing $m$ iterations is $\Pr\left[  N=m\right]  $,
it follows that
\[
\mathrm{MDT}\left(  \mu\right)  =%
{\displaystyle\sum\limits_{m=1}^{\infty}}
\Pr\left[  N=m\right]  \tau^{\intercal}\left(
{\displaystyle\sum\limits_{n=1}^{m}}
M^{n-1}\right)  \mu=\tau^{\intercal}\left(
{\displaystyle\sum\limits_{m=1}^{\infty}}
\Pr\left[  N=m\right]  \left(
{\displaystyle\sum\limits_{n=1}^{m}}
M^{n-1}\right)  \right)  \mu
\]
This proves (\ref{eq:uesta}), as for (\ref{eq:altra}), given any two integers
$n,m\in\mathbb{N}$ set%
\[
1\left(  n\leq m\right)  =\left\{
\begin{array}
[c]{ccc}%
1 &  & \text{if }n\leq m\\
0 &  & \text{if }n>m
\end{array}
\right.
\]
with this%
\begin{align*}%
{\displaystyle\sum\limits_{m=1}^{\infty}}
\Pr\left[  N=m\right]  \left(
{\displaystyle\sum\limits_{n=1}^{m}}
M^{n-1}\right)   &  =%
{\displaystyle\sum\limits_{m=1}^{\infty}}
{\displaystyle\sum\limits_{n=1}^{m}}
\Pr\left[  N=m\right]  M^{n-1}=%
{\displaystyle\sum\limits_{m=1}^{\infty}}
{\displaystyle\sum\limits_{n=1}^{\infty}}
1\left(  n\leq m\right)  \Pr\left[  N=m\right]  M^{n-1}\\
&  =%
{\displaystyle\sum\limits_{n=1}^{\infty}}
\left(
{\displaystyle\sum\limits_{m=1}^{\infty}}
1\left(  n\leq m\right)  \Pr\left[  N=m\right]  \right)  M^{n-1}=%
{\displaystyle\sum\limits_{n=1}^{\infty}}
\Pr\left[  N\geq n\right]  M^{n-1}%
\end{align*}
as wanted.\hfill$\blacksquare$

\subsection{Kolmogorov criterion, Luce product rule, and \textbf{Lemma
\ref{lem:full}}}

Let $A$ be a finite set, with typical elements $i$, $j$ and $k$. A\emph{
(left) stochastic matrix }$P=\left[  P\left(  i\mid j\right)  \right]
_{i,j\in A}$ is an $A\times A$ matrix such that $P\left(  \cdot\mid j\right)
\in\Delta\left(  A\right)  $ for all $j\in A$.\ In general, $P\left(  i\mid
j\right)  $ is interpreted as the probability with which a system moves from
state $j$ to state $i$.

\begin{definition}
\label{def:One}A stochastic matrix $P$ is:

\begin{itemize}
\item \emph{reversible} if there exists $\pi\in\Delta\left(  A\right)  $ such
that
\begin{equation}
P\left(  i\mid j\right)  \pi\left(  j\right)  =P\left(  j\mid i\right)
\pi\left(  i\right)  \qquad\forall i,j\in A \label{eq:reversi}%
\end{equation}

\item \emph{transitive} if
\begin{equation}
P\left(  j\mid i\right)  P\left(  k\mid j\right)  P\left(  i\mid k\right)
=P\left(  k\mid i\right)  P\left(  j\mid k\right)  P\left(  i\mid j\right)
\qquad\forall i,j,k\in A \label{eq:kolmo}%
\end{equation}

\item \emph{full} if $P\left(  i\mid j\right)  >0$ for all $i,j\in A$, i.e.,
$P\left(  \cdot\mid j\right)  \in\Delta_{+}\left(  A\right)  $ for all $j\in
A$;

\item an \emph{exploration matrix} if
\[
P\left(  j\mid i\right)  =P\left(  i\mid j\right)  >0\qquad\forall i\neq j\in
A
\]

\end{itemize}
\end{definition}

A few remarks are in order. First, since (\ref{eq:reversi}) is automatically
satisfied when $i=j$, reversibility can be stated as $P\left(  i\mid j\right)
\pi\left(  j\right)  =P\left(  j\mid i\right)  \pi\left(  i\right)  $ for all
$i\neq j$. Second, transitivity is known as the \emph{Kolmogorov criterion} in
the Markov chains literature (see, e.g., Kelly, 1979, p. 24, and Kijima, 1997,
p. 60) and as the \emph{product rule} in stochastic choice literature (Luce
and Suppes, 1965, p. 341) where it was introduced by Luce (1957).

Third, transitivity is automatically satisfied if at least two of the three
states $i$, $j$, and $k$ in $A$ coincide. In fact,

\begin{itemize}
\item if $i=j$, then
\begin{align*}
P\left(  j\mid i\right)  P\left(  k\mid j\right)  P\left(  i\mid k\right)   &
=P\left(  i\mid i\right)  P\left(  k\mid i\right)  P\left(  i\mid k\right) \\
P\left(  k\mid i\right)  P\left(  j\mid k\right)  P\left(  i\mid j\right)   &
=P\left(  k\mid i\right)  P\left(  i\mid k\right)  P\left(  i\mid i\right)
\end{align*}

\item if $i=k$, then
\begin{align*}
P\left(  j\mid i\right)  P\left(  k\mid j\right)  P\left(  i\mid k\right)   &
=P\left(  j\mid i\right)  P\left(  i\mid j\right)  P\left(  i\mid i\right) \\
P\left(  k\mid i\right)  P\left(  j\mid k\right)  P\left(  i\mid j\right)   &
=P\left(  i\mid i\right)  P\left(  j\mid i\right)  P\left(  i\mid j\right)
\end{align*}

\item if $j=k$, then
\begin{align*}
P\left(  j\mid i\right)  P\left(  k\mid j\right)  P\left(  i\mid k\right)   &
=P\left(  j\mid i\right)  P\left(  j\mid j\right)  P\left(  i\mid j\right) \\
P\left(  k\mid i\right)  P\left(  j\mid k\right)  P\left(  i\mid j\right)   &
=P\left(  j\mid i\right)  P\left(  j\mid j\right)  P\left(  i\mid j\right)
\end{align*}

\end{itemize}

\noindent Therefore, transitivity can be restated as
\[
P\left(  j\mid i\right)  P\left(  k\mid j\right)  P\left(  i\mid k\right)
=P\left(  k\mid i\right)  P\left(  j\mid k\right)  P\left(  i\mid j\right)
\]
for all distinct$\ i$, $j$ and $k$ in $A$.

\begin{boxer}
\textbf{Nota Bene} the argument we just reported applies to any function
$P:A\times A\rightarrow\mathbb{R}$, and it is independent on the values
$P\left(  i\mid i\right)  $ that the function takes on the diagonal.
\end{boxer}

The next result, which relates reversibility and transitivity, builds upon
Kolmogorov (1936) and Luce and Suppes (1965).

\begin{proposition}
\label{prop:kolmo}Let $P$ be a full stochastic matrix. The following
conditions are equivalent:

\begin{enumerate}
\item[(i)] $P$ is reversible, with respect to some $\pi\in\Delta\left(
A\right)  $;

\item[(ii)] $P$ is transitive.
\end{enumerate}

In this case, given any $i\in A$, it holds%
\[
\pi\left(  j\right)  =\frac{\dfrac{P\left(  j\mid i\right)  }{P\left(  i\mid
j\right)  }}{%
{\displaystyle\sum\limits_{k\in A}}
\dfrac{P\left(  k\mid i\right)  }{P\left(  i\mid k\right)  }}\qquad\forall
j\in A
\]
In particular, $\pi$ is unique and has full support.
\end{proposition}

\noindent\textbf{Proof} If $P$ is reversible with respect to $\pi$, then
\begin{equation}
P\left(  i\mid j\right)  \pi\left(  j\right)  =P\left(  j\mid i\right)
\pi\left(  i\right)  \qquad\forall i,j\in A \label{eq:heart}%
\end{equation}
If $\pi\left(  i^{\ast}\right)  =0$ for some $i^{\ast}\in A$, then (since $P$
is full)%
\begin{equation}
\pi\left(  j\right)  =\frac{P\left(  j\mid i^{\ast}\right)  }{P\left(
i^{\ast}\mid j\right)  }\pi\left(  i^{\ast}\right)  =0\qquad\forall j\in A
\label{eq:star}%
\end{equation}
But, this is impossible since $%
{\displaystyle\sum\limits_{j\in A}}
\pi\left(  j\right)  =1$. Hence, $\pi$ has full support. Moreover, by
(\ref{eq:star}),%
\[
\frac{\dfrac{P\left(  j\mid i^{\ast}\right)  }{P\left(  i^{\ast}\mid j\right)
}}{%
{\displaystyle\sum\limits_{k\in A}}
\dfrac{P\left(  k\mid i^{\ast}\right)  }{P\left(  i^{\ast}\mid k\right)  }%
}=\frac{\dfrac{P\left(  j\mid i^{\ast}\right)  }{P\left(  i^{\ast}\mid
j\right)  }\pi\left(  i^{\ast}\right)  }{%
{\displaystyle\sum\limits_{k\in A}}
\dfrac{P\left(  k\mid i^{\ast}\right)  }{P\left(  i^{\ast}\mid k\right)  }%
\pi\left(  i^{\ast}\right)  }=\frac{\pi\left(  j\right)  }{%
{\displaystyle\sum\limits_{k\in A}}
\pi\left(  k\right)  }=\pi\left(  j\right)  \qquad\forall j\in A
\]
irrespective of the choice of $i^{\ast}\in A$. Finally, given any $i,j,k\in
A$, by (\ref{eq:heart}) we have:%
\begin{align*}
\frac{\pi\left(  j\right)  }{\pi\left(  i\right)  }\frac{\pi\left(  k\right)
}{\pi\left(  j\right)  }\frac{\pi\left(  i\right)  }{\pi\left(  k\right)  }
&  =1\implies\frac{P\left(  j\mid i\right)  }{P\left(  i\mid j\right)  }%
\frac{P\left(  k\mid j\right)  }{P\left(  j\mid k\right)  }\frac{P\left(
i\mid k\right)  }{P\left(  k\mid i\right)  }=1\implies\\
\frac{P\left(  j\mid i\right)  P\left(  k\mid j\right)  P\left(  i\mid
k\right)  }{P\left(  k\mid i\right)  P\left(  j\mid k\right)  P\left(  i\mid
j\right)  }  &  =1\implies P\left(  j\mid i\right)  P\left(  k\mid j\right)
P\left(  i\mid k\right)  =P\left(  k\mid i\right)  P\left(  j\mid k\right)
P\left(  i\mid j\right)
\end{align*}
and transitivity holds.

Conversely, if transitivity holds, choose arbitrarily $i^{\ast}\in A$, and set%
\[
\pi^{\ast}\left(  j\right)  \doteqdot\frac{\dfrac{P\left(  j\mid i^{\ast
}\right)  }{P\left(  i^{\ast}\mid j\right)  }}{%
{\displaystyle\sum\limits_{k\in A}}
\dfrac{P\left(  k\mid i^{\ast}\right)  }{P\left(  i^{\ast}\mid k\right)  }%
}\doteqdot\zeta\dfrac{P\left(  j\mid i^{\ast}\right)  }{P\left(  i^{\ast}\mid
j\right)  }\qquad\forall j\in A
\]
where $1/\zeta=%
{\displaystyle\sum\limits_{k\in A}}
P\left(  k\mid i^{\ast}\right)  /P\left(  i^{\ast}\mid k\right)  >0$. With
this, for all $i,j\in A$,
\[
P\left(  i\mid j\right)  \pi^{\ast}\left(  j\right)  =P\left(  i\mid j\right)
\zeta\dfrac{P\left(  j\mid i^{\ast}\right)  }{P\left(  i^{\ast}\mid j\right)
}%
\]
Transitivity implies that%
\[
P\left(  j\mid i\right)  P\left(  i^{\ast}\mid j\right)  P\left(  i\mid
i^{\ast}\right)  =P\left(  i^{\ast}\mid i\right)  P\left(  j\mid i^{\ast
}\right)  P\left(  i\mid j\right)
\]
and since $P$ is full%
\[
P\left(  j\mid i\right)  \frac{P\left(  i\mid i^{\ast}\right)  }{P\left(
i^{\ast}\mid i\right)  }=\frac{P\left(  j\mid i^{\ast}\right)  }{P\left(
i^{\ast}\mid j\right)  }P\left(  i\mid j\right)  \qquad\forall j\in A
\]
Thus,%
\[
P\left(  i\mid j\right)  \zeta\frac{P\left(  j\mid i^{\ast}\right)  }{P\left(
i^{\ast}\mid j\right)  }=P\left(  j\mid i\right)  \zeta\frac{P\left(  i\mid
i^{\ast}\right)  }{P\left(  i^{\ast}\mid i\right)  }=P\left(  j\mid i\right)
\pi^{\ast}\left(  i\right)
\]
and reversibility with respect to $\pi^{\ast}$ holds.\hfill$\blacksquare
\bigskip$

\noindent\textbf{Proof of Lemma \ref{lem:full}} We have $M\left(  \cdot\mid
j\right)  \in\Delta_{+}\left(  A\right)  $ for all $j\in A$. Indeed,%
\[
M\left(  i\mid j\right)  =\underset{\in\left(  0,1\right]  }{~\underbrace
{Q\left(  i\mid j\right)  }}\times\underset{\in\left(  0,1\right)
}{\underbrace{\rho\left(  i\mid j\right)  }}~\in\left(  0,1\right)
\qquad\forall i\neq j
\]
and $M\left(  j\mid j\right)  =1-%
{\displaystyle\sum\limits_{k\neq j}}
Q\left(  k\mid j\right)  \rho\left(  k\mid j\right)  >1-%
{\displaystyle\sum\limits_{k\neq j}}
Q\left(  k\mid j\right)  =Q\left(  j\mid j\right)  \geq0$. \hfill
$\blacksquare$

\subsection{Proof of Proposition \ref{prop:trans}}

\noindent\textbf{Proof} For clarity, we split point (i) in two parts:

\begin{itemize}
\item[(i.a)] the transition matrix $M$ is reversible (for every exploration
matrix $Q$);

\item[(i.b)] the transition matrix $M$ is reversible (for some exploration
matrix $Q$);
\end{itemize}

We show that (i.a) $\implies$ (i.b) $\implies$ (ii) $\implies$ (iii)
$\implies$ (i.a).

(i.a) $\implies$ (i.b) Trivial.

(i.b) $\implies$ (ii) By Proposition \ref{prop:kolmo}, $M$ is transitive.
Thus, for all distinct $i$, $j$ and $k$ in $A$,%
\begin{align*}
&  Q\left(  j\mid i\right)  \rho\left(  j\mid i\right)  Q\left(  k\mid
j\right)  \rho\left(  k\mid j\right)  Q\left(  i\mid k\right)  \rho\left(
i\mid k\right) \\
&  =Q\left(  k\mid i\right)  \rho\left(  k\mid i\right)  Q\left(  j\mid
k\right)  \rho\left(  j\mid k\right)  Q\left(  i\mid j\right)  \rho\left(
i\mid j\right)
\end{align*}
Since $Q$ is symmetric and strictly positive off the diagonal, then
\[
\rho\left(  j\mid i\right)  \rho\left(  k\mid j\right)  \rho\left(  i\mid
k\right)  =\rho\left(  k\mid i\right)  \rho\left(  j\mid k\right)  \rho\left(
i\mid j\right)
\]
that is, $\rho$ is transitive.

(ii)$\ \implies\ $(iii) Arbitrarily extend $\rho$ to $A^{2}$, say by taking
$\rho\left(  i\mid i\right)  =1$ for all $i\in A$. Choose $i^{\ast}\in A$, and
set%
\[
\pi^{\ast}\left(  j\right)  \doteqdot\frac{\dfrac{\rho\left(  j\mid i^{\ast
}\right)  }{\rho\left(  i^{\ast}\mid j\right)  }}{%
{\displaystyle\sum\limits_{k\in A}}
\dfrac{\rho\left(  k\mid i^{\ast}\right)  }{\rho\left(  i^{\ast}\mid k\right)
}}\qquad\forall j\in A
\]
Note that
\[
\pi^{\ast}\left(  i^{\ast}\right)  \doteqdot\frac{1}{1+%
{\displaystyle\sum\limits_{k\neq i^{\ast}}}
\dfrac{\rho\left(  k\mid i^{\ast}\right)  }{\rho\left(  i^{\ast}\mid k\right)
}}%
\]
does not depend on the value of $\rho\left(  i^{\ast}\mid i^{\ast}\right)  $.
With this%
\[
\frac{\pi^{\ast}\left(  j\right)  }{\pi^{\ast}\left(  k\right)  }%
\doteqdot\frac{\dfrac{\rho\left(  j\mid i^{\ast}\right)  }{\rho\left(
i^{\ast}\mid j\right)  }}{\dfrac{\rho\left(  k\mid i^{\ast}\right)  }%
{\rho\left(  i^{\ast}\mid k\right)  }}\qquad\forall j,k\in A
\]
The product rule implies, again irrespective of the choice of $\rho\left(
h\mid h\right)  $ for $h\in A$,%
\[
\rho\left(  j\mid i\right)  \rho\left(  k\mid j\right)  \rho\left(  i\mid
k\right)  =\rho\left(  k\mid i\right)  \rho\left(  j\mid k\right)  \rho\left(
i\mid j\right)  \qquad\forall i,j,k\in A
\]
and%
\[
\frac{\rho\left(  j\mid i\right)  }{\rho\left(  i\mid j\right)  }\frac
{\rho\left(  i\mid k\right)  }{\rho\left(  k\mid i\right)  }=\frac{\rho\left(
j\mid k\right)  }{\rho\left(  k\mid j\right)  }%
\]
Therefore,%
\begin{equation}
\frac{\pi^{\ast}\left(  j\right)  }{\pi^{\ast}\left(  k\right)  }%
\doteqdot\dfrac{\rho\left(  j\mid i^{\ast}\right)  }{\rho\left(  i^{\ast}\mid
j\right)  }\dfrac{\rho\left(  i^{\ast}\mid k\right)  }{\rho\left(  k\mid
i^{\ast}\right)  }=\frac{\rho\left(  j\mid k\right)  }{\rho\left(  k\mid
j\right)  }\qquad\forall j,k\in A \label{eq:sticazzi}%
\end{equation}
Then,%
\[
\frac{\pi^{\ast}\left(  j\right)  }{\pi^{\ast}\left(  j\right)  +\pi^{\ast
}\left(  k\right)  }=\frac{\rho\left(  j\mid k\right)  }{\rho\left(  j\mid
k\right)  +\rho\left(  k\mid j\right)  }\qquad\forall j\neq k\text{ in}\ A
\]
and%
\[
\rho\left(  j\mid k\right)  =\frac{\pi^{\ast}\left(  j\right)  }{\pi^{\ast
}\left(  j\right)  +\pi^{\ast}\left(  k\right)  }\,\underset{\doteqdot
s\left(  j,k\right)  }{\underbrace{(\rho\left(  j\mid k\right)  +\rho\left(
k\mid j\right)  )}}%
\]
This shows that (\ref{eq:Haste}) holds. In this case, $M$ is reversible with
respect to $\pi^{\ast}$ since, for all $k\neq j$ in $A$,%
\begin{equation}
M\left(  k\mid j\right)  \pi^{\ast}\left(  j\right)  =Q\left(  k\mid j\right)
\underset{=\rho\left(  j\mid k\right)  \pi^{\ast}\left(  k\right)  \text{ by
(\ref{eq:sticazzi})}}{\underbrace{\rho\left(  k\mid j\right)  \pi^{\ast
}\left(  j\right)  }}=Q\left(  j\mid k\right)  \rho\left(  j\mid k\right)
\pi^{\ast}\left(  k\right)  =M\left(  j\mid k\right)  \pi^{\ast}\left(
k\right)  \label{eq:stella}%
\end{equation}
irrespective of $Q$. This is the basic idea behind Hastings (1970).

(iii)$\ \implies\ $(i.a) If there exist $\pi\in\Delta_{+}\left(  A\right)  $
and a symmetric $s:A_{\neq}^{2}\rightarrow\left(  0,\infty\right)  $ such
that, for all $k\neq j\ $in $A$,%
\[
\rho\left(  k\mid j\right)  =s\left(  k,j\right)  \dfrac{\pi\left(  k\right)
}{\pi\left(  k\right)  +\pi\left(  j\right)  }%
\]
then for all $k\neq j$ in $A$,%
\begin{align*}
M\left(  k\mid j\right)  \pi\left(  j\right)   &  =Q\left(  k\mid j\right)
\rho\left(  k\mid j\right)  \pi\left(  j\right)  =Q\left(  k\mid j\right)
s\left(  k,j\right)  \dfrac{\pi\left(  k\right)  }{\pi\left(  k\right)
+\pi\left(  j\right)  }\pi\left(  j\right) \\
&  =Q\left(  j\mid k\right)  s\left(  j,k\right)  \dfrac{\pi\left(  j\right)
}{\pi\left(  j\right)  +\pi\left(  k\right)  }\pi\left(  k\right)  =Q\left(
j\mid k\right)  \rho\left(  j\mid k\right)  \pi\left(  k\right)  =M\left(
j\mid k\right)  \pi\left(  k\right)
\end{align*}
irrespective of $Q$. Then, $M$ is reversible with respect to $\pi$ for all
$Q$, which proves (11.a).

Moreover, if also $\tilde{\pi}\in\Delta_{+}\left(  A\right)  $ and $\tilde
{s}:A_{\neq}^{2}\rightarrow\left(  0,\infty\right)  $ are such that
(\ref{eq:Haste}) holds, the argument we just presented implies that $M$ is
also reversible with respect to $\tilde{\pi}$ for all $Q$. Proposition
\ref{prop:kolmo} tells us that $\pi=\tilde{\pi}$, which not only proves
uniqueness of $\pi$, but also, thanks to (\ref{eq:stella}), that $\pi
=\pi^{\ast}$ implying (\ref{eq:expli}).

Uniqueness of $s$ follows by inverting (\ref{eq:Haste}). In fact, since $\pi$
is unique and $\rho=\left\{  \rho\left(  i\mid j\right)  \right\}  _{i\neq j}$
is given, then (\ref{eq:Haste}) implies that
\[
s\left(  i,j\right)  =\dfrac{\pi\left(  i\right)  +\pi\left(  j\right)  }%
{\pi\left(  i\right)  }\rho\left(  i\mid j\right)  \qquad\forall\left(
i,j\right)  \in A_{\neq}^{2}%
\]
identifying $s$.

Another application of (\ref{eq:Haste}), yields
\[
\frac{\pi\left(  j\right)  }{\pi\left(  k\right)  }=\frac{\dfrac{\pi\left(
j\right)  }{\pi\left(  j\right)  +\pi\left(  k\right)  }s\left(  j,k\right)
}{\dfrac{\pi\left(  k\right)  }{\pi\left(  k\right)  +\pi\left(  j\right)
}s\left(  k,j\right)  }=\frac{\rho\left(  j\mid k\right)  }{\rho\left(  k\mid
j\right)  }\qquad\forall j,k\in A
\]
Thus, if $\rho$ is unbiased,%
\begin{equation}
\frac{\pi\left(  j\right)  }{\pi\left(  j\right)  +\pi\left(  k\right)
}=\frac{\rho\left(  j\mid k\right)  }{\rho\left(  j\mid k\right)  +\rho\left(
k\mid j\right)  }=\frac{\rho\left(  j\mid k\right)  }{1}\qquad\forall j,k\in A
\label{eq:ue}%
\end{equation}
and
\[
s\left(  i,j\right)  =\underset{=\frac{1}{\rho\left(  i\mid j\right)  }\text{
by (\ref{eq:ue})}}{\underbrace{\dfrac{\pi\left(  i\right)  +\pi\left(
j\right)  }{\pi\left(  i\right)  }}}\rho\left(  i\mid j\right)  =1
\]
and $s\equiv1$. Conversely, if $s\equiv1$, then (\ref{eq:Haste}) implies that
$\rho$ is unbiased.\hfill$\blacksquare$

\hfill\textit{Milan, \today}

\end{document}